\documentclass[conference,9pt]{IEEEtran}
\IEEEoverridecommandlockouts
\usepackage{cite}
\usepackage{amsmath,amssymb,amsfonts}
\usepackage{algorithmic}
\usepackage{graphicx}
\usepackage{textcomp}
\usepackage{xcolor}
\usepackage{multirow}
\usepackage{float}
\usepackage{booktabs}
\def\BibTeX{{\rm B\kern-.05em{\sc i\kern-.025em b}\kern-.08em
    T\kern-.1667em\lower.7ex\hbox{E}\kern-.125emX}}
\begin{document}

\title{
Capturing Rich Behavior Representations: A Dynamic Action Semantic-Aware Graph Transformer for Video Captioning
\thanks{Supported by Civil Aviation Joint Fund of National Natural Science Foundation of China under Grant U2333206 and 2024 Tianjin 'the Belt and Road' Joint Laboratory Project under the Special Innovation Platform Program of Tianjin Science and Technology Plan: China Kazakhstan Joint Research Center for Digital Twin Airport of Civil Aviation University of China(24PTLYHZ00230).}
}
\author{Caihua Liu\textsuperscript{1,2}, Xu Li\textsuperscript{1,2}, Wenjing Xue\textsuperscript{1,2}, Wei Tang\textsuperscript{1,2}, Xia Feng\textsuperscript{1,2,3,*}\thanks{*corresponding Author:xfeng@cauc.edu.cn.}\\
College of Computer Science and Technology, Civil Aviation University of China,Tianjin, China\textsuperscript{1}\\
Key Laboratory of Smart Airport Theory and System, CAAC, 2898 Jinbei Road, Dongli District, Tianjin, China\textsuperscript{2}\\
Science and Technology Innovation Research Institute,Civil Aviation University of China, Tianjin\textsuperscript{3}}
\maketitle

\begin{abstract}
Existing video captioning methods merely provide shallow or simplistic representations of object behaviors, resulting in superficial and ambiguous descriptions. However, object behavior is dynamic and complex. To comprehensively capture the essence of object behavior, we propose a dynamic action semantic-aware graph transformer. Firstly, a multi-scale temporal modeling module is designed to flexibly learn long and short-term latent action features. It not only acquires latent action features across time scales, but also considers local latent action details, enhancing the coherence and sensitiveness of latent action representations. Secondly, a visual-action semantic aware module is proposed to adaptively capture semantic representations related to object behavior, enhancing the richness and accurateness of action representations. By harnessing the collaborative efforts of these two modules, we can acquire rich behavior representations to generate human-like natural descriptions. Finally, this rich behavior representations and object representations are used to construct a temporal objects-action graph, which is fed into the graph transformer to model the complex temporal dependencies between objects and actions. To avoid adding complexity in the inference phase, the behavioral knowledge of objects is distilled into a simple network through knowledge distillation. The experimental results on MSVD and MSR-VTT datasets demonstrate that the proposed method achieves significant performance improvements across multiple metrics.
\end{abstract}


\section{Introduction}
Video captioning aims to automatically generate a sentence described in natural language for video. It has gained extensive attention from researchers due to its potential applications, such as video retrieval\cite{yu2017end}, human-robot interaction\cite{nguyen2018translating}, and visual question answering\cite{gordon2018iqa,wu2017visual}.\\
\begin{figure}[h] 
\centering 
\begin{minipage}{\textwidth} 
\includegraphics[width=8.8cm, height=1.5cm]{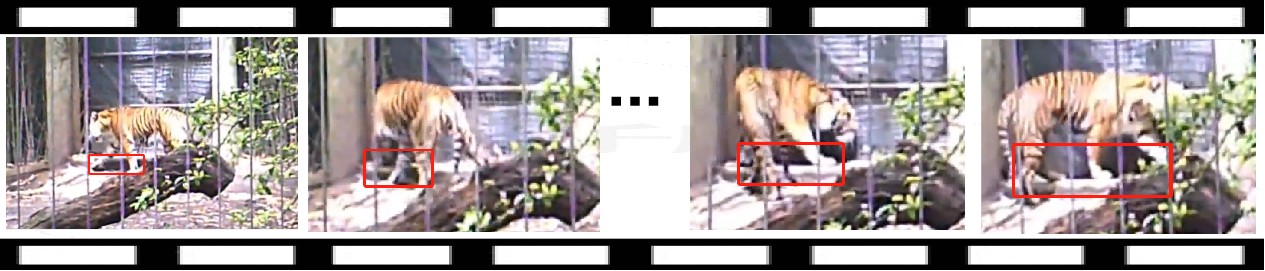}  
\\ \hspace*{1em} (a) GT: A tiger is \textcolor{green}{ walking around} in an enclosure.
\\ \hspace*{2.4em}  Baseline: A tiger is \textcolor{red}{ running}.
\label{fig:Introduction3.1}  
\end{minipage}
\begin{minipage}{\textwidth}  
\includegraphics[width=8.7cm,height=1.5cm]{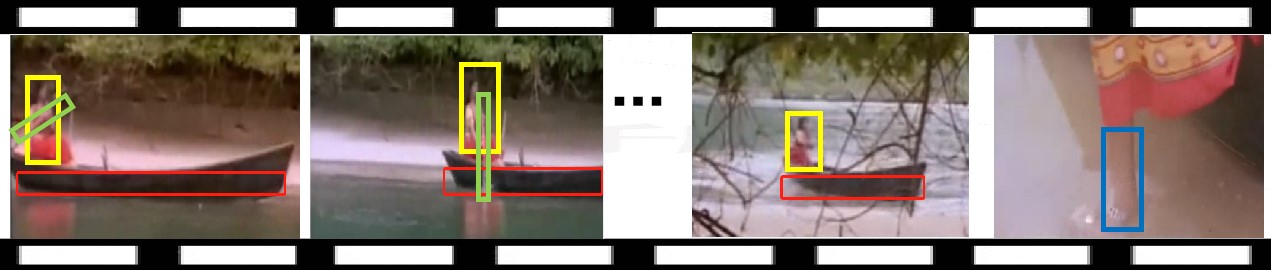} 
\\ \hspace*{1em} (b) GT: A woman \textcolor{green}{ paddles} a canoe and then \textcolor{green}{disembarks}. 
\\ \hspace*{2.4em} Baseline: A woman is \textcolor{red}{riding} a boat. 
\\
\label{fig:Introduction3.3}  
\end{minipage}  
\\
\caption{Qualitative examples of captioning that do not adequately capture the essence of the object's behavior.}  
\vspace{-10pt}
\label{fig:total}  
\end{figure}
\indent The key to captioning video is to clearly describe what objects perform what behaviors in what scenes. Existing video captioning methods \cite{sutskever2014sequence,yao2015describing,venugopalan2014translating}are mainly devoted to the feature representations of objects and scenes with 2DCNN\cite{chua1997cnn} and 3DCNN\cite{6165309}. Partial methods\cite{ye2022hierarchical,qu2020ksf,zhang2019object} focus on capturing key objects. Recently, researchers introduce spatio-temporal relational graph \cite{xue2022exploring,pan2020spatio,zhang2020object}to represent the relationship between objects. However, they ignore the behavioral semantics of objects, leading to simple behavioral descriptions. Moreover, some multimodal large models \cite{zhang2023video,lin2023video,li2023videochat}attempt to understand video contents through visual question answering. Although they are good at comprehending the attributes of objects, their understanding of object behaviors remains insufficient.\\
\begin{figure*}[h]
\centering
\includegraphics[width=\textwidth]{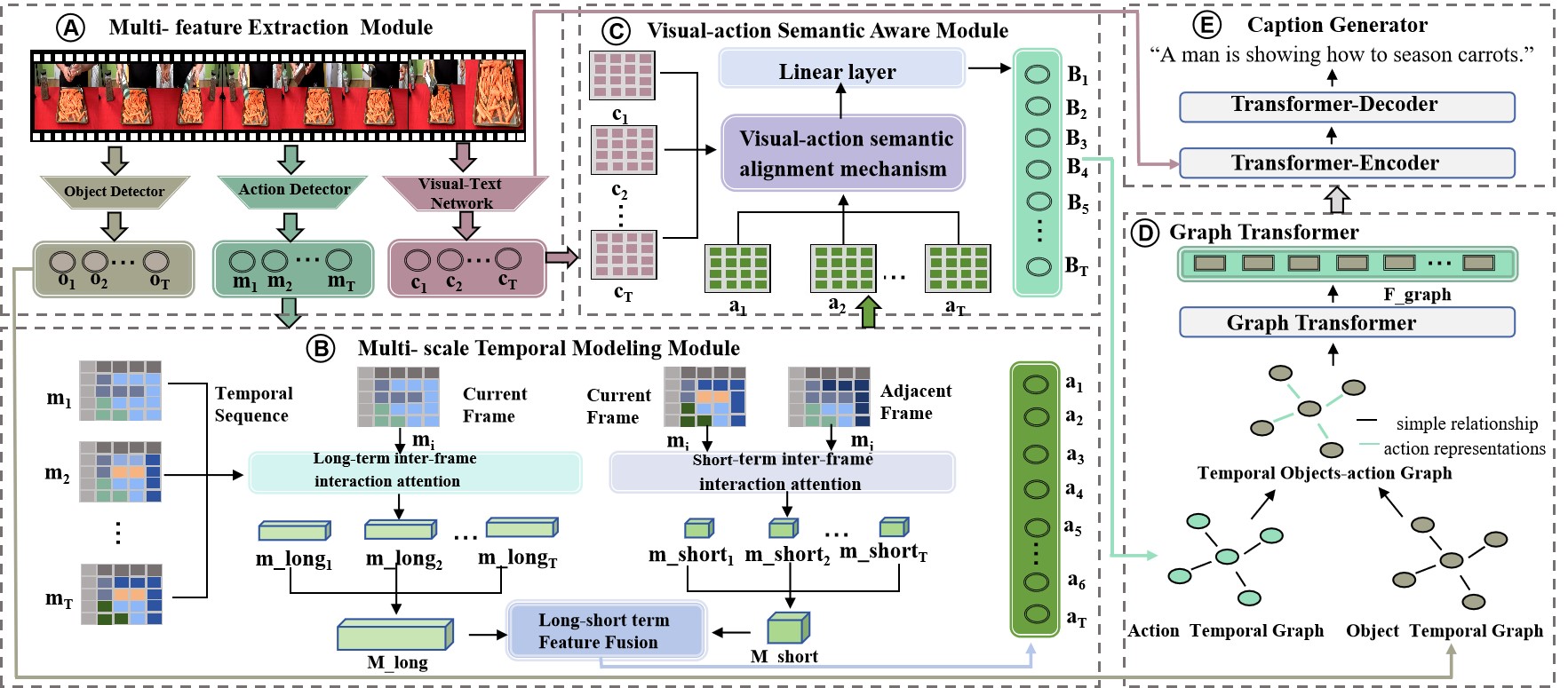}
\caption{The pipleline of the proposed method. For training, the proposed graph transformer network and the visual-text network are trained simultaneously. In the model inference phase, only the visual-text network is used because the visual-text network has already learned the object behavior knowledge of the whole network through the knowledge distillation process. This approach avoids complex computations and increases the speed of inference.}
\label{fig:pipline14}
\end{figure*}
\indent Object behavior is dynamic and complex, defined by a series of interrelated latent actions. Existing methods\cite{ye2022hierarchical}\cite{zhai2023joint} still have two shortcomings in understanding the semantic behavior of objects. \textbf{(a)Ignoring latent actions temporal variations.} The long-short variations of latent actions may be an indispensable clue to the understanding of object behavior. As shown in part (a) of Figure 1, when attention is dominated by short-term keyframes, it makes the model hallucinate and generate incorrect description "a tiger is running". If the model can balance its attention between long-term variation and the details of short-term keyframes, the actual behavior "walking around" can be captured. As shown in part (b) of Figure 1, if long-term variations of latent actions play a decisive role like "paddling a canoe", short-term variations of latent actions like "disembark" may be overlooked, leading to the generation of incomplete descriptions like "riding a boat". Thus, the video captioning method needs to dynamically adjust long and short-term latent action features for a more accurate and comprehensive behavior description. \textbf{(b)Simplifying the semantic description of object behavior.} Existing methods are based on I3D\cite{carreira2017quo} and C3D\cite{tran2015learning} to obtain the object's action information, resulting in inaccurate or simple generation of object behaviors, such as "running" and "riding". If semantic representations that are more closely related to object behavior can be captured with the help of powerful visual-text representations, then the model will generate more accurate and semantically deep descriptions of object behavior.\\
\indent To solve these problems, we propose a dynamic action semantic-aware graph transformer. A multi-scale temporal modeling module is designed, which can flexibly learn both long and short-term latent action features. The dynamic fusion of long and short-term latent action features improves the coherence and sensitivity of latent action representations. Meanwhile, Considering the complexity of object behaviors, a visual-action semantic aware module is designed, which adaptively selects and learns the representations that are closely related to the semantics of object behaviors, enhancing the richness and accurateness of action representations.\\
\indent In summary, our main contributions are as follows:
\begin{itemize}
    \item
    We propose a multi-scale temporal modeling module that can flexibly learn long and short-term latent action features. It not only capture latent action tendencies across time scales, but also consider local latent action details, enhancing the coherence and sensitiveness of latent action representations.
    \item A visual-action semantic aware module is devised, which can adaptively capture  the representations closely related to behavioral semantics. The rich action representations help the model's understanding of behavior, enhancing the richness and accurateness of action representations.
    \item Experimental results show that our method achieves a significant improvement over the state-of-the-art on MSVD and MSR-VTT datasets.
\end{itemize}
\section{METHODOLOGY}
As shown in Figure 2, the proposed method can be divided into five modules.The multi-feature extraction module is used to extract various features. The multi-scale temporal modeling module could be used to flexibly learn long and short-term latent action features. The visual-action semantic aware module is designed to learn the representations closely relevant to the semantics of object behaviors. The graph transformer is introduced to link object representations with behavior representations. The caption generation model uses visual features to generate text descriptions.
\subsection{Multi-feature Extraction Module}
 Following the feature extraction setting of Hendria et al\cite{hendria2023action}, we extract object features and action features, denoted as $O$=\{$o_1$, $o_2$, ..., $o_T$\} and $M$=\{$m_1$,$m_2$, ..., $m_T$\}. The purpose of extracting object features is to obtain comprehensive information about the object and to construct object temporal graph. Action features are used to explore the essence of behavior. To learn the semantic information related to the behavior, we introduce  Clip4clip\cite{luo2022clip4clip} model as visual-text network. The Visual-text features extracted  from Clip4clip are denoted as $C$=\{$c_1$,$c_2$, ..., $c_T$\}.
\begin{table*}
 \caption {Performance comparison with the state-of-the-art on MSVD and MSR-VTT. "*" indicates the reproduced method. The best results are shown in bold."-" means the number not available. JKSUC represents the jounal named J KING SAUD UNIV-COM.}
\centering
\resizebox{\linewidth}{!}{%
\begin{tabular}{cc|cccc|cccc}
\hline 
\multirow{2}{*}{Methods} & \multirow{2}{*}{Venue} & 
\multicolumn{4}{c|}{MSVD} & \multicolumn{4}{c}{MSR-VTT} \\
\cline{3-6} \cline{7-10}
 & & BLEU-4 & METEOR & ROUGE-L & CIDEr & BLEU-4 & METEOR & ROUGE-L & CIDEr \\ 
\hline OA-BTG\cite{zhang2019object}& CVPR '2019 & 56.9 & 36.2 & - & 90.6 & 41.4 & 28.2 & - & 46.9\\ 
\hline ORG-TRL\cite{zhang2020object}& CVPR '2020 & 54.3 & 36.4 & 73.9 & 95.2 & 43.6 & 28.8 & 62.1 & 50.9\\ 
\hline MGRMP\cite{chen2021motion} & ICCV '2021 & 55.8 & 36.9 & 74.5 & 98.5 & 41.7 & 28.9 & 62.1 & 51.4 \\ 
\hline O2NA\cite{liu2021o2na} & ICCV '2021 &  55.4 & 37.4 & 74.5 & 96.4 & 41.6 & 28.5 & 62.4 & 51.1\\ 
\hline CLIP4Caption\cite{tang2021clip4caption}& MM' 2021 & - & - & - & - & 46.1 & 30.7 & 63.7 & 57.7 \\ 
\hline LSTG \cite{li2022long}& TIP'2022 & 55.6 & 37.1 & 73.5 & 98.5 & 42.6 & 28.3 & 61.0 & 49.5 \\ 
\hline HMN\cite{ye2022hierarchical} & CVPR'2022 & 59.2 & 37.7 & 75.1 & 104.0 & 43.5 & 29.0 & 62.7 & 51.5 \\ 
\hline SwinBERT\cite{lin2022swinbert} & CVPR'2022 & 58.2 & 41.3 & 77.5 & 120.6 & 45.4 & 30.6 & 64.1 & 55.9\\ 
\hline LWCG\cite{verma2023leveraging}& AAAI '2023 & 60.9 & 38.2 & 75.3 & 92.11 & 46.1 & 29.9 & 63.8 & 53.7 \\ 
\hline Clip4VideoCap\cite{mahmud2023clip4videocap}& ICASSP '2023 & - & - & - & - & - & 31.5 & 65.8 & 62,2 \\ 
\hline SEM-POS\cite{nadeem2023sem} & CVPR'2023 & 60.1 & 38.5 & 76.0 & 108.3 & 45.2 & 30.7 & 64.1 & 53.1 \\ 
\hline AKVA-grid\cite{hendria2023action} & JKSUC'2023 & \textbf{62.90} & 41.81 & \textbf{78.80} & 119.07 & 49.10 & 31.57 & 65.52 & 61.27 \\ 
\hline AKVA-object\cite{hendria2023action}* &JKSUC'2023 & 61.33 & 41.24 & 78.60 & 118.14 & 48.86 & 31.97 & 65.95 & 61.27 \\ 
\hline ours & & 62.45 & \textbf{41.92} & 78.69 & \textbf{121.24} & \textbf{50.05} & \textbf{32.43} & \textbf{66.49} & \textbf{63.25} 
\\
\hline
\end{tabular}
}   
\end{table*}
\subsection{Multi-scale Temporal Modeling Module}
Multi-scale Temporal Modeling Module consists of three sub components. The long-term inter-frames interaction attention aims to model the latent action dependencies of each frame and the whole sequence, improving the coherence of latent action representations. The short-term inter-frames interaction attention focus on latent action interactions between adjacent frames, enhancing the sensitiveness of latent action representations. Long-short term features fusion enables focus on long-term actions while remaining sensitive to new actions, and also maintains attention on short-term actions without losing touch with long-term tendencies.\\
\textbf{Long-term inter-frames interaction attention.} For the action feature sequence $M$=\{$m_1$,$m_2$,...,$m_T$\}, we design a attention mechanism to learn the relationship between $m_i$ and $M$. Three learnable parameters $W^{long}_q$, $W^{long}_k$ and $W^{long}_v$, are introduced to help the model capture important information within the features.
 For each $m_i$ and $M$, we parameterise the representation as $Q_i$ , $K$ and $V$ as follows: \begin{equation}
Q_i = m_i \cdot W^{long}_q, \quad K = M \cdot W^{long}_k, \quad V = M \cdot W^{long}_v
\end{equation} where $i$ is the index of current frame.\\
\indent The scores and weights of attention are expressed through equations 2 and 3, respectively.
\begin{equation}
\text{Attention\_scores}_{\text{long}}[i] = Q_i \cdot K^T
\end{equation}
\begin{equation}
\text{Attention\_weights}_{\text{long}}[i] = \text{softmax}\left(\frac{\text{Attention\_scores}_{\text{long}}[i]}{\sqrt{\text{scale}}}\right)
\end{equation}
\indent Then, we dynamically capture the latent behavioral trends by aggregating the long-term interactions of each frame through weighted summation, as shown in equation 4:
\begin{equation}\text{M\_long}[i] = \sum_{i=1}^{T} \text{Attention\_weights}_{\text{long}}[i] \times V\end{equation} 
\indent Finally, the long-term action representations of the whole video are denoted as ${M\_long}$=\{$m\_long_1$,$m\-long_2$,...$m\_long_T$\}, $m\_long_i$ represents the short-term action representations of each moment.\\
\textbf{Short-term inter-frames interaction attention.} Referring to the long-term approach, a attention mechanism is designed to dynamically learn the short-term action relationship between $m_i$ and $m_j$ as follows:\begin{equation}
Q_i = m_i \cdot W^{short}_q, \quad K_j = m_j \cdot W^{short}_k, \quad V_j = m_j \cdot W^{short}_v
\end{equation} 
where $i$ is the index of current frame, $j$ is the index of adjacent frames.
\begin{equation}\text{Cross\_scores}_{\text{short}}[i][j] = Q_i \times K_j^T\end{equation}
\begin{equation}
\text{Cross}_{\text{short}}[i][j] = \text{softmax} \left( \frac{\text{Cross\_scores}_{\text{short}}[i][j]}{\sqrt{\text{scale}}} \right)\end{equation}
\begin{equation}
\text{M\_short}[i] = \sum_{i=1}^{j} \text{Cross}_{\text{short}}[i][j] \times \text{V}_j\end{equation}
\indent Finally, we obtain the short-term action representations of the whole video as ${M\_short}$=\{$m\_short_1$,$m\_short_2$,...$m\_short_T$\}.\\
\textbf{Long-short term Features Fusion.} Long-term features ${M\_long}$ and short-term features ${M\_short}$ are concatenated to get long-short term features ${A\_fused}$=\{$a_1$,$a_2$,...$a_T$\}. The fused features consider the changes in action features at different time scales and re-represent the action features, resulting in more comprehensive representations of actions.
\subsection{Visual-action Semantic Aware Module} 
\indent To further enhance the semantic expression ability of action features, a visual-action semantic aware module is proposed. This module aligns long-short term action features ${A\_fused}$ with visual-text features $C$ (mentioned in section A). This module enabling the model to comprehend the semantic correlation between the visual features and textual descriptions, generating relevant semantic representations.\\
\indent In order to adaptively learn the most relevant information from the visual-textual features to the action features, we introduce three learnable parameters, $W^{vt}_q$, $W^{vt}_k$ and $W^{vt}_v$. $Q_i$, $K_i$ and $V_i$ as computed as follows:
\begin{equation}
Q_i = c_i\cdot W^{vt}_q, \quad K_i = a_i \cdot W^{vt}_k, \quad V_i = a_i \cdot W^{vt}_v
\end{equation}
where $i$ is the index of each frame.

\indent The weights of attention are expressed through equations 10, representing the correlation between the fused action features and visual-text features. .
\begin{equation}
    \text{Attention\_weights} = \text{softmax}(\frac{Q_i \times K_i^T}{\sqrt{\text{scale}}})
\end{equation} 
\indent Specifically, higher attention weights indicate a stronger association between the fused action features and visual-text features. We utilize attention weights to perform a weighted sum of the visual-text features, denoted as B:
\begin{equation}
\text{B} = \sum_{i=1}^{T} (\text{Attention\_weights}_i \times \text{V}_i)\end{equation}
\subsection{Graph Transformer}
 Inspired by the method of Pan et al\cite{pan2020spatio}, we construct an object temporal graph to track variations of objects across video frames, by calculating the cosine similarity of object features in adjacent frames. Following the work of Hendria et al\cite{hendria2023action}, a new action temporal graph is constructed, which is based on the visual representations closely relevant to behavior semantics. Using the object temporal graph and the action temporal graph, a novel temporal objects-action graph is designed to link object representations and action representations. The structure of the graph transformer\cite{yun2019graph} allows for a more detailed and accurate representations of the relationship between objects and actions over time. Thus, the temporal objects-action graph is fed into the graph transformer, resulting in a new tensor representation of object-action features, denoted as $F_{graph}$.
\subsection{Caption Generator} 
\indent We simultaneously train two networks through knowledge distillation\cite{hendria2023action}. For training, the visual-text network along with $F_{graph}$ are fed into the Transformer. a cross-entropy loss function\cite{zhang2020object} is employed to constrain the distance between the generated captions and the ground truth. Kullback-Leibler\cite{pan2020spatio} is  also introduced to measure the difference between the probability distributions of words generated by these two networks, thus enabling the visual-text network to learn from the proposed graph transformer network.
\section{EXPERIMENTS}
\subsection{Datasets, Evaluation metrics and Experimental Details}\label{AA}
The experiments are conducted on two widely used benchmark datasets, MSVD\cite{pei2019memory} and MSR-VTT\cite{wang2019controllable}. We employ the following evaluation metrics: BLEU@4\cite{papineni2002bleu}, METEOR\cite{banerjee2005meteor}, ROUGE-L\cite{lin2004rouge}, CIDEr\cite{vedantam2015cider}. BLEU@4 (here n = 4). Specifically, CIDEr is a metric designed specifically for captioning tasks that places more emphasis on the semantic consistency of video descriptions. The learning rates of MSVD and MSR-VTT are set to 1e-4 and 3e-4 respectively. The number of epochs are set as 60 and 80 for MSVD and  MSR-VTT datasets. The video captioning model is trained on RTX 3090.
\subsection{Quantitative Analysis}
The proposed algorithm is compared with object-reated methods\cite{zhang2019object}\cite{zhang2020object}\cite{liu2021o2na}, temporal-related methods\cite{chen2021motion}\cite{li2022long}, multimodal-related methods\cite{verma2023leveraging}\cite{tang2021clip4caption}\cite{mahmud2023clip4videocap}\cite{nadeem2023sem} and other methods\cite{ye2022hierarchical}\cite{lin2022swinbert}. The experimental results show that the proposed method outperforms the comparison methods. This means that video captioning can be significantly improved by exploring the temporal dependencies of objects and actions. Wherein AKVA-object\cite{hendria2023action} is our direct baseline, which also establishes a link between objects and actions. However, it doesn't consider the dynamic and semantic complexity of object behavior. AKVA-grid\cite{hendria2023action} is slightly higher on BLUE-4 and ROUGE-L on the small MSVD dataset. The main reason is that object behaviors in MSVD are relatively simple and easily learned. On the other hand, AKVA-grid uses grid features rather than object features, which introduces more visual information. Notably, our model achieves state-of-the-art performance on large-scale MSR-VTT dataset. One reason is larger dataset contain a greater number of videos, the model can learn richer visual features. Another reason is both the object’s behaviors and descriptions in MSR-VTT dataset are complex. It is difficult to generate accurate descriptions without exploring temporal action relations and object behavior semantics. The above analysis shows that capturing rich behavior representations can improve the accurateness. 

\begin{table}[H]
\caption{Ablation study on the MSR-VTT dataset in terms of BLEU-4, METEOR, ROUGE-l, and CIDEr scores.}
\centering
\begin{tabular}{ccccc}
\toprule
Models& BLEU-4 & METEOR & ROUGE-L&CIDEr \\ \hline
baseline*&48.86&31.97&65.95 &61.27\\
w/o visual-{action} &49.97&32.29&66.05 &62.98\\
w/o temporal&49.88&32.33&66.42&62.28\\
full model&\textbf{50.05}&\textbf{32.43}&\textbf{66.49} &\textbf{63.25}\\
\bottomrule
\end{tabular}
\end{table}
\subsection{Ablation Study}
To validate the effectiveness of the proposed method, we carried out ablation experiments on MSR-VTT dataset as shown in Table 2. The first row is the baseline method, the fourth row is the full model proposed in this paper. The result of the proposed method without the multi-scale temporal modeling module is shown in the second row, and the proposed method without visual-action semantic aware module is shown in the third row. Compared to the baseline model, the performance of the second and third rows improves $1.84 \sim 2.84$ on CIDEr. It is proved that the temporal information of actions and visual representations related to behaviour semantics play an important role in improving the quality of video captioning. The full model achieves 3.1 improvement on CIDEr, which suggests these two modules can promote each other to get a better result. The ablation experiment of our method on MSVD dataset is consistent with that of MSR-VTT.
\begin{figure}[h] 
\centering 
\begin{minipage}{\textwidth} 
\includegraphics[width=8.8cm]{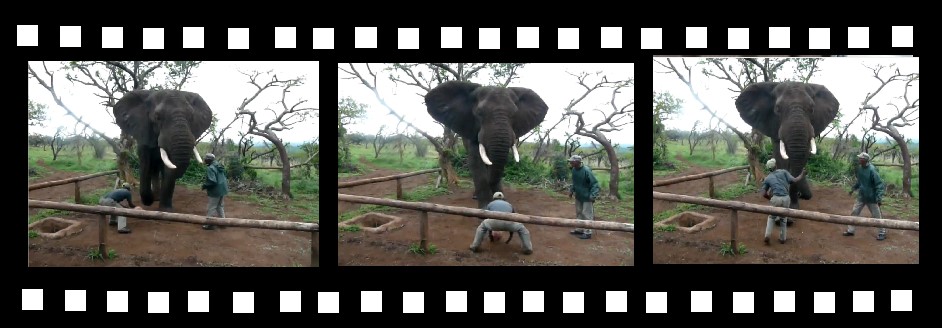}  
\\ GT:Two trainer are training the elephant in the jungle.
\\ Baseline: A man is \textcolor{red}{ pushing} a rhino.
\\ Ours: Two person are \textcolor{green}{  training}the elephant.
\label{fig:icassp1}  
\end{minipage}
\begin{minipage}{\textwidth}  
\includegraphics[width=8.7cm]{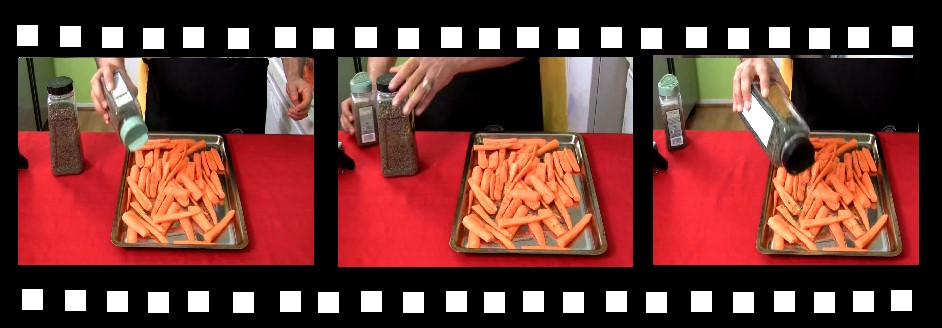} 
\\ GT:A man is demonstrating how to season carrots.
\\ Baseline: A man \textcolor{red}{pours} different bottles. 
\\ Ours:A man is \textcolor{green}{showing} how to season carrots. 
\\
\label{fig:icassp2}  
\end{minipage}  
\\
\caption{Visualization examples for qualitative comparisons between our
method and the baseline model (better viewed in color).}  
\vspace{-10pt}
\label{fig:dingxing}  
\end{figure}
\subsection{Qualitative Analysis}
Two case studies are shown to illustrate the quality of the proposed method compared with the baseline model. For the first example, the baseline model generates "a man is pushing a rhino". Compared to ground truth, the baseline model not only misidentifies objects in the video, but also misrepresents the video content. Our method combines the action variations of objects in the temporal sequence and generates "two persons are training the elephant". For the second example, the baseline model generates "a man pours different bottles". Compared to ground truth, the baseline model focuses on the action between the person and the bottle, and does not have a deeper understanding of the dynamic latent action variations, leading to a shallow description. Our method explores the deeper meanings behind the actions, generating "a man is showing how to season carrots". The above experimental results show that the proposed method outperforms the baseline on the accurateness and comprehensiveness of descriptions\\
\section{CONCLUSION}\label{SCM}
In this paper, we propose a dynamic action semantic-aware graph transformer to improve coarse behavioral representations. To comprehensively understand the essence of object behavior, we design two main modules. The first module is a multi-scale temporal modeling module that can flexibly learn long and short-term latent action features. Besides, this module mainly consists of long-term inter-frames interaction attention and short-term inter-frames interaction attention. The long-term inter-frames interactive attention aims to establish long-term dependence of action features. The short-term inter-frames interaction attention aims to establish short-term dependence. The second module is a visual-action semantic aware module that can adaptively capture representations closely related to behavioral semantics. Finally, we distill knowledge into a simple network through an object action graph transformer model. Extensive experiments validate the validity of our method.
\clearpage
\bibliographystyle{IEEEtran}
\bibliography{myfile}
\end{document}